\documentclass[conference]{IEEEtran}
\IEEEoverridecommandlockouts
\usepackage{cite}
\usepackage{amsmath,amssymb,amsfonts}
\usepackage{algorithmic}
\usepackage{graphicx}
\usepackage{multirow}
\usepackage{booktabs}
\usepackage{url}
\usepackage{textcomp}
\usepackage{xcolor}
\def\BibTeX{{\rm B\kern-.05em{\sc i\kern-.025em b}\kern-.08em
    T\kern-.1667em\lower.7ex\hbox{E}\kern-.125emX}}
\begin{document}

\title{Compound Density Networks for Risk Prediction using Electronic Health Records \\
}

\author{
    \IEEEauthorblockN{
        Yuxi Liu\thanks{Yuxi Liu is the corresponding author.}\IEEEauthorrefmark{1},
        Shaowen Qin\IEEEauthorrefmark{1},
        Zhenhao Zhang\IEEEauthorrefmark{2},
        Wei Shao\IEEEauthorrefmark{3},
    }
    \IEEEauthorblockA{
        \IEEEauthorrefmark{1}College of Science and Engineering, Flinders University, Adelaide, SA, Australia\\
        \IEEEauthorrefmark{2}College of Life Sciences, Northwest A\&F University, Yangling, Shaanxi, China\\
        \IEEEauthorrefmark{3}College of Electrical and Computer Engineering, UC Davis, Davis California, CA, USA\\
            \{liu1356, shaowen.qin\}@flinders.edu.au
            zhangzhenhow@nwafu.edu.cn
            weishao@ucdavis.edu
    }
}
\maketitle

\begin{abstract}
Electronic Health Records (EHRs) exhibit a high amount of missing data due to variations of patient conditions and treatment needs. Imputation of missing values has been considered an effective approach to deal with this challenge. Existing work separates imputation method and prediction model as two independent parts of an EHR-based machine learning system. We propose an integrated end-to-end approach by utilizing a Compound Density Network (\textit{CDNet}) that allows the imputation method and prediction model to be tuned together within a single framework. \textit{CDNet} consists of a Gated recurrent unit (GRU), a Mixture Density Network (MDN), and a Regularized Attention Network (RAN). The GRU is used as a latent variable model to model EHR data. The MDN is designed to sample latent variables generated by GRU. The RAN serves as a regularizer for less reliable imputed values. The architecture of \textit{CDNet} enables GRU and MDN to iteratively leverage the output of each other to impute missing values, leading to a more accurate and robust prediction. We validate \textit{CDNet} on the mortality prediction task on the MIMIC-III dataset. Our model outperforms state-of-the-art models by significant margins. We also empirically show that regularizing imputed values is a key factor for superior prediction performance. Analysis of prediction uncertainty shows that our model can capture both aleatoric and epistemic uncertainties, which offers model users a better understanding of the model results.
\end{abstract}

\begin{IEEEkeywords}
Electronic Health Records, missing data imputation, model uncertainty, machine learning, data mining.
\end{IEEEkeywords}

\section{Introduction}
The immense accumulation of Electronic Health Records (EHRs) provides an unprecedented opportunity to develop accurate, meaningful, and reliable outcome prediction models \cite{dusenberry2020analyzing, zhang2020inprem, chen2021unite}.

However, health data in EHRs present a high degree of irregularity, due to variations of patient conditions and treatment needs. One of the notable issues is missing values, which makes accurate and reliable predictions challenging. We present an example of a patient's records from the MIMIC-III dataset \cite{johnson2016mimic} in Fig. 1. The physician conducts/prescribes the necessary lab tests each time a patient is seen. Different physiological variables (e.g., heart rate, glucose) are examined at different times depending on the patient's symptoms \cite{tan2020data}. When certain symptom disappears, corresponding variables are no longer examined. This results in missing values.
\begin{figure}[!htb]
        \centering
        \includegraphics[width = 0.9\linewidth]{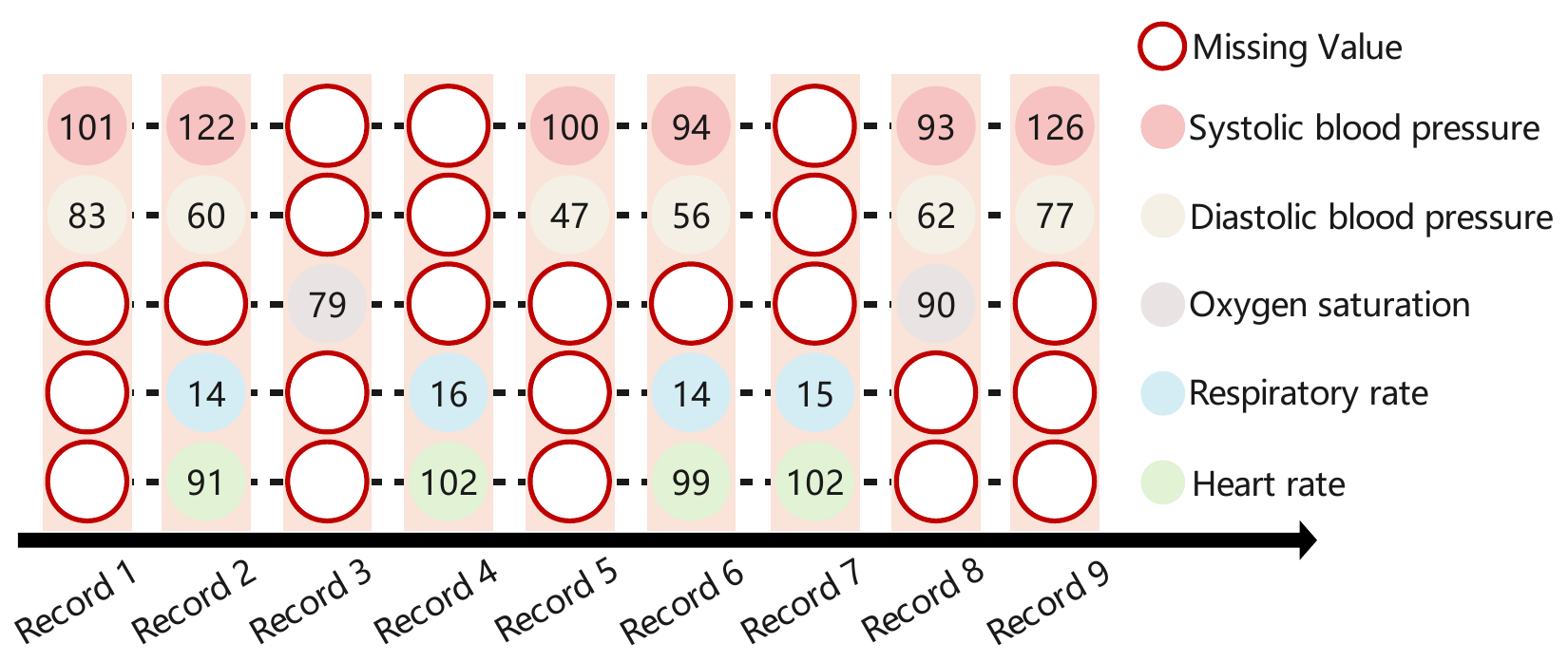}
        \caption{\textbf{An example of a patient's clinical records.}}
        \label{fig:DataFrame}
\end{figure}

Machine learning algorithms have brought revolutionary changes in a wide variety of fields, such as computer vision \cite{khan2020machine}, machine translation \cite{singh2017machine}, and computational biology \cite{tarca2007machine}. Machine learning algorithms build models based on a large amount of sample data, known as training data. The prediction performance of machine learning models will suffer if the training dataset has a large number of missing values. A simple solution is to remove the observations that have missing data. This, however, is not applicable to EHR data, as it means valuable information is discarded. A better strategy would be to impute missing values. Several methods currently exist for the imputation of missing values, such as Multiple Imputation, Expectation-Maximization, Nearest Neighbor and Hot Deck methods. These methods rely on variable correlations to impute missing values. Previous studies of \cite{bertsimas2021imputation, welch2014evaluation, bounthavong2015approach} have demonstrated the effectiveness of these methods on EHR data.

Existing work usually separates imputation method and prediction model as two independent parts of an EHR-based machine learning system. After imputing missing values, the complete EHR data matrix is fed into existing machine learning models to make risk predictions. For example, machine learning models can be used to predict in-hospital mortality, decompensation, length-of-stay, and phenotype classification \cite{harutyunyan2019multitask, sheikhalishahi2019benchmarking}.

However, with an EHR dataset, caution must be taken, as the missing values are likely attributed to patient symptoms (as mentioned earlier). For this reason, the imputation method and prediction model should be tuned together within a single framework rather than separated as two parts. By doing so, the prediction model is able to deal with the missing values in EHR data effectively. A recent study by \cite{che2018recurrent} developed a deep prediction model named GRU-D to address this problem. The overall structure of GRU-D is built upon Gated recurrent units (GRU) \cite{cho2014learning}. The GRU-D mainly incorporates the empirical mean value and the previous observation to impute missing values. Despite its efficacy, the GRU-D suffers from notable methodological weaknesses. The empirical mean value might be biased due to the diversity of patient data, hence lacking reliability. Utilizing less reliable imputed values as part of the input is equivalent to introducing noise/error into the input. This inevitably introduces a high degree of aleatoric uncertainty into the dataset (e.g., imputation error). Most of the existing machine learning models are sensitive to aleatoric uncertainty, that is, a few small variations in the inputs may lead to significant changes in model outputs. This might lead to biased prediction due to the cumulative effect of imputation errors.

Uncertainty quantification has a pivotal role in model optimization \cite{abdar2021review}. Uncertainty quantification allows researchers to know how confident they can be with the prediction results, which is essential to building trust in prediction models. In contrast, it is often less trustworthy when prediction results are presented without uncertainty quantification. There are two main types of uncertainties: epistemic and aleatoric \cite{hullermeier2021aleatoric}. Epistemic uncertainty indicates what the model does not know. It is attributed to inadequate knowledge of the model. This is the uncertainty that can be reduced by having more data. Aleatoric uncertainty is the inherent uncertainty that is part of the data generating process, such as sensor noise, record error, or missing value. This variability cannot be reduced by having more data. Current studies have used these two uncertainties as indicators of the reliability of the method \cite{li2016aleatory, zhang2020inprem}.

The study by \cite{kendall2017uncertainties} investigated the differential impact of aleatoric uncertainty and epistemic uncertainty on computer vision tasks. The approach taken in this study is a mixed methodology based on Bayesian theory and deep neural networks, known as Bayesian neural networks (BNNs). The core idea of BNNs is to replace the deterministic network's weight parameters with their probability distributions and, instead of optimizing the network weights directly, use the average of all possible weights. However, the inference of BNNs remains a major challenge and incurs a huge computational cost \cite{rohekar2018constructing}. To address the issue, many variational inference techniques are proposed, such as stochastic variational inference \cite{hoffman2013stochastic} and sampling-based variational inference \cite{domke2018importance},  which have achieved promising performance in many prediction tasks \cite{dhaka2020robust, acerbi2018variational}. Despite their efficacy, these methods still impose a tremendous burden on computational costs.

One well-known study that is often cited in research on Bayesian inference approximation is that of \cite{gal2016dropout}, which found that the use of dropout in deep neural networks could be regarded as an approximate Gaussian process. Their theoretical framework employs a dropout layer as a Bayesian inference approximation before every weight layer. The use of dropout as a Bayesian approximation is currently the most popular method for providing epistemic uncertainty estimation due to its low computation cost and high efficiency. Despite this, the use of dropout requires a number of repeated feed-forward calculations of deep neural networks with randomly sampled weight parameters. The resulting outputs are used to quantify the epistemic uncertainty of those deep neural networks.

In this paper, we propose an end-to-end, novel, and robust prediction model by utilizing a Compound Density Network (\textit{CDNet}) that consists of a Gated recurrent unit (GRU), a Mixture Density Network (MDN), and a Regularized Attention Network (RAN). The proposed \textit{CDNet} enables GRU and MDN to work together by iteratively leveraging the output of each other to impute missing values. The GRU is used as a latent variable model to model EHR data. The MDN is designed to sample latent variables generated by GRU. The sampling process is equivalent to exploiting the generated latent variables to model the distribution of imputed features. The MDN is built from two components: a deep neural network and a mixture of distributions. We assume the mixture of distributions comprises multiple Gaussian distributions because the imputed features are continuous. Specifically, latent variables are fed into the deep neural network. The deep neural network then provides the parameters for multiple Gaussian distributions, including their means, variances, and weights that can be used to build a Gaussian mixture distribution. The resulting Gaussian mixture distribution is a multimodal distribution that contributes to the modeling of complex patterns found in the input.

To enhance the reliability of imputed values and quantify their uncertainties, the RAN is served as a regularizer for less reliable imputed values, leading to more robust model outputs. The core idea behind RAN design is to model the attention weights as a function of the variance of Gaussian mixture distribution. When used for regularized learning, it assigns smaller weights to imputed values with large variance. The output of RAN is fed into the developed predictor network to make risk predictions. This involves making an MDN for predicting the class probability distribution. The modeling process of the MDN includes learning about the impact of aleatoric uncertainty and epistemic uncertainty. When used for quantifying epistemic uncertainty, MDN can be regarded as a sampling-free method because it does not require repeated feed-forward calculations of deep neural networks. Specifically, the MDN uses a deep neural network to provide the parameters (i.e., mean and variance) for a mixture of distributions. When properly trained, we obtain the mean and the standard deviation, which means we have the entire class probability distribution (e.g., the risk of death and no death) and, by extension, the estimate of the aleatoric and epistemic uncertainty. The resulting predicted class probability distributions are further used to estimate risk probabilities (e.g., the probability of death).

We validate \textit{CDNet} on the mortality prediction task from a publicly available EHR dataset that has a large number of missing values in the input. Our model outperforms state-of-the-art models by significant margins. We also empirically show that regularizing imputed values is a key step for superior prediction performance. Analysis of prediction uncertainty shows that our model can capture both aleatoric and epistemic uncertainties, which allows model users to gain a better understanding of the model results.

\section{Methods}
We describe the proposed \textit{CDNet} in this section. We introduce the basic notations of risk prediction tasks first. We then detail the \textit{CDNet} architecture. Finally, we present how to use \textit{CDNet} for risk prediction tasks.
\subsection{Basic Notations}
The EHR data consists of patients' time-ordered records. Each patient's records ensemble can be further categorized as a patient journey, termed EHR patient journey data in the following sections. The EHR patient journey data is denoted by $X^{p}$ = $[X_{1}^{p}, \cdots, X_{t}^{p}, \cdots, X_{T_{p}}^{p}]$ $\in$ $\mathbb{R}^{N \times T_{p}}$, where $N$ is the number of sequential dynamic features (that occur over time, e.g., vital signs) and $T_{p}$ is the number of records. For simplicity, we drop the $p$ when it is unambiguous in the following sections.
\subsection{\textit{CDNet} architecture}
\subsubsection{Learning feature embedding}
An essential step before implementing the proposed components is to learn the embedding of sequential dynamic features. Learning feature embeddings is able to help us translate feature spaces. Specifically, an embedding layer is applied to sequential dynamic features, generating the corresponding representations. This embedding layer provides a mapping between sequential dynamic features and embedding space, allowing GRU to learn the underlying dynamics of patient journeys via lower-dimensional feature representations.

Let $Z$ denotes learnable feature vectors, which are used as prefilled values of the patient journey $X$. This $Z$ is initialized from the standard Gaussian distribution. By doing this, the $X$ is now termed $X^{\prime}$. Given an input $X^{\prime}$, the embedding layer can be written as:
\begin{equation}
\begin{split}
\label{eq:1}
X^{emb} = W^{emb} \cdot X^{\prime} + b^{emb}
\end{split}
\end{equation}
where $X^{emb} \in \mathbb{R}^{d_{emb} \times T}$ is the learned sequential dynamic feature embedding. $W^{emb} \in \mathbb{R}^{d_{emb} \times N}$ is a learnable parameter, $b^{emb} \in \mathbb{R}^{d_{emb}}$ is a bias, and $d_{emb}$ is the dimension of the embedding space.

\subsubsection{Model components}
\paragraph{Gated recurrent units (GRU)}
The core idea of GRU is to exploit the degree of missingness of all EHR patient journeys to impute missing values. Due to patient-specific symptoms, the degree of missingness of sequential dynamic features may vary among patient journeys. Based on this assumption, missing values of one patient journey can be inferred from other EHR patient journeys. The inference process is achieved by employing GRU \cite{cho2014learning}. The GRU is currently the most popular method for generating latent variables from multivariate time series data. Latent variables are a transformation of the data points into a continuous lower-dimensional space. EHR patient journey data is a type of multivariate time series data with more than one time-dependent feature; each not only depends on its past values but also has some dependency on others. These dependencies must be modeled, which are used for forecasting future values. After training, the employed GRU is able to generate a series of latent variables derived from all EHR patient journeys modeling. These latent variables correspond one-to-one with sequential dynamic features.

GRU is a variant of Recurrent neural networks (RNN) that modifies the basic RNN's hidden layer. One advantage of the GRU is that it avoids the problem of the vanishing gradient suffered by an RNN. The essential nature of GRU is the gating of the hidden state. Given input $X_{t}^{emb} \in \mathbb{R}^{d_{emb}}$ and previous hidden state $H_{t - 1} \in \mathbb{R}^{g}$, the current hidden state $H_{t}$ can be obtained through the following steps.

Specifically, $X^{emb}_{t}$ and $H_{t-1}$ are fed into a gating mechanism. The gating mechanism, including a reset gate $R_{t}$ and an update gate $U_{t}$, is to decide which of the previous information will be retained for $H_{t}$. The objective function of the gating mechanism can be written as:
\begin{equation}
\begin{split}
\label{eq:2}
R_{t} = \sigma (W_{R} \cdot [H_{t - 1}, X^{emb}_{t}] + b_{R}) \\
U_{t} = \sigma (W_{U} \cdot [H_{t - 1}, X^{emb}_{t}] + b_{U})
\end{split}
\end{equation}
where $W_{R} \in \mathbb{R}^{g \times (d_{emb} + g)}$ and $W_{U} \in \mathbb{R}^{g \times (d_{emb} + g)}$ are learnable parameters. $b_{R} \in \mathbb{R}^{g}$ and $b_{U} \in \mathbb{R}^{g}$ are biases. $\sigma$ is the sigmoid activation function that is used to normalize the outputs $R_{t}$ and $U_{t}$ in $[0, 1]$. The $X_{t}^{emb}$ and the element-wise multiplication of $H_{t-1}$ with $R_{t}$ are used to generate an intermediate $\widetilde{H}_{t}$. $H_{t}$ is obtained by the element-wise convex combinations between $\widetilde{H}_{t}$ and $U_{t}$. The formula can be written as:
\begin{equation}
\begin{split}
\label{eq:3}
\widetilde{H}_{t} = tanh (W_{H} \cdot [R_{t} \odot H_{t - 1}, X^{emb}_{t}] + h_{H}) \\
H_{t} = U_{t} \odot H_{t - 1} + (1 - U_{t}) \odot \widetilde{H}_{t}
\end{split}
\end{equation}
where $W_{H} \in \mathbb{R}^{g \times (d_{emb} + g)}$ is a learnable parameter and $h_{H} \in \mathbb{R}^{g}$ is a bias. $\odot$ denotes the element-wise multiplication.

The generated latent variables $\{H_{t}\}^{T}_{t=1}$ are used as the input of MDN.

\paragraph{Mixture Density Network (MDN)}
The MDN is designed to sample latent variables generated by GRU. The sampling process is equivalent to exploiting the use of generated latent variables to model the distribution of imputed features. The MDN comprises a deep neural network and a mixture of distributions. Since the imputed features are continuous,  we assume the mixture of distributions comprises multiple Gaussian distributions. Specifically, latent variables are fed into the deep neural network. The deep neural network then provides the parameters for multiple Gaussian distributions, including their means and variances, as well as weights that can be used to build a Gaussian mixture distribution. The Gaussian mixture distribution can be written as:
\begin{equation}
\begin{split}
\label{eq:4}
p(X_{t} | H_{t}) = \sum^{K}_{k = 1} \beta_{k} \cdot D_{k} (X_{t} | H_{t}) \\
D_{k} (X_{t} | H_{t}) = N(\mu_{k}, \varSigma_{k})
\end{split}
\end{equation}
where $k$ denotes the index of the corresponding mixture component. There are up to $K$ mixture components (i.e., distributions) per output. $\beta$ denotes the mixing parameter. $D$ is the corresponding distribution to be mixed. $D$ is a multivariate Gaussian distribution, where $\mu$ is the mean vector and $\varSigma$ is the covariance matrix with $\sigma^{2}$ on the diagonal and 0 otherwise.

We assume mean $\mu$ and variance $\sigma^{2}$ of each distribution are derived from a nonlinear combination of the inputs. A deep feed-forward network is modified to output the parameters of the Gaussian mixture distribution. A constraint we must enforce here is $\sigma^{2}$ $>$ 0, i.e., the variance of Gaussian must be positive. This is implemented by employing the Exponential Linear Unit (ELU) activation with an offset \cite{guillaumes2017mixture}. Since this can technically be zero, we have added an $\epsilon$ to the modified ELU to ensure stability.
\begin{equation}
\begin{split}
\label{eq:5}
h = ReLU(W^{h} \cdot H + b^{h}) \\
\beta = softmax(W^{\beta} \cdot h + b^{\beta}) \\
\mu_{k} = W_{k}^{\mu} \cdot h + b^{\mu}_{k} \\
\sigma^{2}_{k} = ELU(W^{\sigma}_{k} \cdot h +b^{\sigma}_{k}) + 1 + \epsilon
\end{split}
\end{equation}
where all parameters of $W$ are projection matrices and all parameters of $b$ are bias vectors. $\epsilon$ is a constant term (e.g., $1 \times 10^{-15}$). $\beta$ is the mixture weight of each component. We use the softmax function to keep $\beta$ in the probability space.

(The sampling process of MDN) We apply Gaussian noise $\xi$ and variance $\varSigma_{k}$ to $\mu_{k}$ to obtain the predicted EHR patient journey $\hat{X}$.
\begin{equation}
\begin{split}
\label{eq:6}
\widetilde{X}_{k} = \mu_{k} + \sqrt{\varSigma_{k}} \cdot \xi, \xi \sim \mathcal{N}(0, 1) \\
\hat{X} = \sum^{K}_{k = 1} \beta_{k} \cdot \widetilde{X}_{k}
\end{split}
\end{equation}

The optimization objective of MDN is to make the predicted patient journey $\hat{X}$ as close to the real-valued patient journey $X$ as possible. The optimization function can be written as:
\begin{equation}
\begin{split}
\label{eq:7}
\mathcal{L} = MSE(W^{proj} \cdot \hat{X}, X^{emb})
\end{split}
\end{equation}
where $MSE(\cdot)$ denotes the mean squared error. $W^{proj} \in \mathbb{R}^{d_{emb} \times N}$ is a learnable projection matrix, which translates the predicted patient journey $\hat{X}$ into the embedding space.

\paragraph{Regularized Attention Network (RAN)}
The output of MDN is a Gaussian mixture distribution. The predicted patient journey $\hat{X}$ is obtained from the sampling of Gaussian mixture distribution. The $\hat{X}$ includes imputed values, combined with real-valued values as a complete data matrix that can be analyzed by our predictor network. However, caution must be taken with imputed values, as they are inferred from the real-valued EHR patient journey data and thus are likely to be less reliable. In response to this issue, the RAN is developed to enhance the reliability of imputed values and quantify their uncertainties. The RAN contains an attention layer; its output is a set of weights. The general idea of the attention layer is to regularize the weights assigned to different patient journeys. For example, it assigns smaller weights to less reliable data.

The unreliability scores of real-valued and imputed values are defined as:
\begin{equation}
\begin{split}
\label{eq:8}
\varphi =
  \begin{cases}
    0,       & for\ real\ valued\ values \\
    \sigma^{2},  & for\ imputed\ values
  \end{cases}
\end{split}
\end{equation}
where $\sigma^{2} = \sum_{k=1}^{K}{\beta_{k} \cdot \sigma_{k}^{2}}$ is the mixed variance of Gaussian mixture distribution that can be used to represent aleatoric uncertainty describing the unreliability of imputed values. Since the real-valued values involve no uncertainty, we set their unreliability scores to zero.

Given the input $\varphi$, the attention layer can be written as:
\begin{equation}
\begin{split}
\label{eq:9}
    \gamma = softmax (W_{\gamma} \cdot (1 - \varphi) + b_{\gamma})
\end{split}
\end{equation}
where $W_{\gamma} \in \mathbb{R}^{N \times N}$ is a learnable parameter and $b_{\gamma} \in \mathbb{R}^{N}$ is a bias.

The weight $\gamma$ is used to regularize the predicted patient journey $\hat{X}$. The formula can be written as:
\begin{equation}
\begin{split}
\label{eq:10}
    \hat{X}^{RAN} = ReLU (W_{RAN} (\gamma \odot \hat{X}) + b_{RAN})
\end{split}
\end{equation}
where $W_{RAN} \in \mathbb{R}^{N \times N}$ is a learnable parameter and $b_{RAN} \in \mathbb{R}^{N}$ is a bias. $\odot$ denotes the element-wise multiplication. $ReLU (\cdot)$ is an activation function.

\subsection{Risk Prediction}
In order to apply \textit{CDNet} to risk prediction tasks, a predictor network is developed, which consists of an attention layer and an MDN.

The $\hat{X}^{RAN}$ (the output of RAN) includes enhanced imputed values, combined with the real-valued patient journey $X$ as a complete data matrix that can be analyzed by the predictor network. The complete data matrix is denoted by $\hat{X}^{Combined}$. Since $\hat{X}^{Combined}$ still takes the form of sequence data, it is difficult to use as the input of an MDN to obtain prediction probability distributions. In response to such an issue, the designed attention layer is used to integrate the $\hat{X}^{Combined}$ into a whole representation. The attention layer can be written as:
\begin{equation}
\begin{split}
\label{eq:11}
\tau = softmax (W_{\tau} \cdot \hat{X}^{Combined} + b_{\tau}) \\
\hat{X}^{Overall} = \sum^{T}_{t = 1} \tau_{t} \odot \hat{X}^{Combined}_{t}
\end{split}
\end{equation}
where $W_{\tau} \in \mathbb{R}^{N \times N}$ and $b_{\tau} \in \mathbb{R}^{N}$ are learnable parameters. $\hat{X}^{Overall}$ is the weighted average of sampling a record according to its importance.

Instead of predicting a deterministic value for each patient journey, we predict the class probability distribution and moreover include aleatoric and epistemic uncertainty estimations. We model the output of every class as an MDN, generating three groups of parameters for every class: the mean $\mu_{p, k}$, the variance $\varSigma_{p, k}$, and the weights of the mixture $\beta_{p}$. The process can be formalized as:
\begin{equation}
\begin{split}
\label{eq:12}
z_{p} = ReLU (W^{z}_{p} \cdot \hat{X}_{p}^{Overall} + b^{z}_{p}) \\
\beta_{p} = softmax (W^{\beta}_{p} \cdot z_{p} + b^{\beta}_{p}) \\
\mu_{p, k} = W^{\mu}_{p, k} \cdot z_{p} + b^{\mu}_{p, k} \\
\sigma^{2}_{p, k} = ELU (W^{\sigma}_{p, k} \cdot z_{p} + b^{\sigma}_{p, k}) + 1 + \epsilon
\end{split}
\end{equation}
where all parameters of $W$ are projection matrices and all parameters of $b$ are bias vectors. $\epsilon$ is a constant term. $p$ denotes the index of the corresponding patient journey. There are up to $P$ patient journeys. $k$ denotes the index of the corresponding mixture component. There are up to $K$ mixture components. $\beta_{p}$ is the mixture weight for each component of patient p's journey. $\mu_{p, k}$ is the predicted value of the k-th component of patient p's journey. $\varSigma_{p, k}$ is the variance for each coordinate $\sigma^{2}_{p, k}$ representing its aleatoric uncertainty. Note that for the binary classification task,
both the dimensions of $\mu_{p, k}$ and $\sigma^{2}_{p, k}$ are set to 2. We use the softmax function to keep $\beta_{p}$ in probability space and use the ELU function again to satisfy the positiveness constraint of the variance.

We apply Gaussian noise $\eta$ and variance $\varSigma_{p, k}$ to $\mu_{p, k}$ to obtain the predicted class probability distribution for patient p's journey.
\begin{equation}
\begin{split}
\label{eq:13}
\tilde{y}_{p, k} =  \mu_{p, k} + \sqrt{\varSigma_{p, k}} \cdot \eta, \eta \sim \mathcal{N}(0, 1) \\
\hat{y}_p = softmax (\sum^{K}_{k = 1} \beta_{p, k} \cdot \tilde{y}_{p,k})
\end{split}
\end{equation}
where $K$ is the number of mixture components. $\hat{y}_p$ is the prediction probability. The objective function $\mathcal{L}^{\prime}$ of the risk prediction task is the average of cross-entropy:
\begin{equation}
\begin{split}
\label{eq:14}
\mathcal{L}^{\prime} = - \frac{1}{P} \sum^{P}_{p = 1} (y_{p}^{\top} \cdot log (\hat{y}_p) + (1 - y_{p})^{\top} \cdot log (1 - \hat{y}_p))
\end{split}
\end{equation}
where $P$ is the number of patient journeys. $y_{p}$ is the ground-truth class/label for patient p's journey.

\section{EXPERIMENTS}
\subsection{Datasets and Tasks}
We conduct the mortality prediction experiments on the publicly available MIMIC-III dataset \cite{johnson2016mimic}. MIMIC-III is one of the largest publicly available ICU datasets, comprising 38,597 distinct patients and a total of 53,423 ICU stays. We use clinical times series data (e.g., heart rate, glucose) as input \cite{harutyunyan2019multitask}. The prediction tasks here are three binary classification tasks: 1) In-hospital mortality (24 hours after ICU admission): to evaluate ICU mortality based on the data from the first 24 hours after ICU admission. 2) In-hospital mortality (36 hours after ICU admission): to evaluate ICU mortality based on the data from the first 36 hours after ICU admission. 3) In-hospital mortality (48 hours after ICU admission): to evaluate ICU mortality based on the data from the first 48 hours after ICU admission. The dataset being used has a high degree of missingness in the input. E.g., for the mortality prediction of the first 48 hours after ICU admission, the results of the statistical analysis of the input are shown in Table I.
\begin{table}[htbp]
  \centering
  \caption{Features from MIMIC-III used in predictions.}
    \begin{tabular}{llr}
    \toprule
    Feature & Type  & \multicolumn{1}{l}{Missingness (\%)} \\
    \midrule
    Capillary refill rate & categorical & 99.78 \\
    Diastolic blood pressure & continuous & 30.90 \\
    Fraction inspired oxygen & continuous & 94.33 \\
    Glascow coma scale eye & categorical & 82.84 \\
    Glascow coma scale motor & categorical & 81.74 \\
    Glascow coma scale total & categorical & 89.16 \\
    Glascow coma scale verbal & categorical & 81.72 \\
    Glucose & continuous & 83.04 \\
    Heart Rate & continuous & 27.43 \\
    Height & continuous & 99.77 \\
    Mean blood pressure & continuous & 31.38 \\
    Oxygen saturation & continuous & 26.86 \\
    Respiratory rate & continuous & 26.80 \\
    Systolic blood pressure & continuous & 30.87 \\
    Temperature & continuous & 78.06 \\
    Weight & continuous & 97.89 \\
    pH    & continuous & 91.56 \\
    \bottomrule
    \end{tabular}%
  \label{tab:addlabel}%
\end{table}%

\subsection{Baselines}
\begin{itemize}
    \item Mean: The mean values of variables are used to impute the missing values.
    \item K-Nearest Neighbor (KNN): The average values of the top $K$ most similar collections are used to impute the missing values.
    \item MICE: Multiple Imputation by Chained Equations (MICE) \cite{van2011mice} uses chain equations to create multiple imputations for variables of different types.
    \item Simple: Simple concatenates the measurement with masking and time intervals, which are then fed into a predictor to make risk predictions \cite{che2018recurrent}.
    \item BRNN: Bidirectional-RNN (BRNN) \cite{suo2019recurrent} generates the imputed values for each variable with the last observed value or the mean values of the same variable. These generated values are used as initial imputed values for the complete data matrix, fed into a bidirectional RNN to predict missing values.
    \item CATSI: CATSI \cite{yin2019context} comprises a context-aware recurrent imputation and a cross-variable imputation, which are used to capture longitudinal information and cross-variable relations from MTS data. A fusion layer in CATSI is used to integrate the two imputation outputs into the final imputations.
    \item BRITS: BRITS \cite{cao2018brits} employs a bidirectional RNN to impute missing values first and then exploits these imputed values to predict the final values.
    \item GRU-D: GRU-D \cite{che2018recurrent} is described in the introduction section. GRU-D also utilized a time decay mechanism to deal with irregular time intervals of medical events in longitudinal patient records. The time decay mechanism builds upon an implicit assumption that the more recent events are more important than previous events on patient-specific risk prediction tasks, hence, taking a monotonically way to decay the information from previous time steps for all patients (the previous medical events).
    \item GRU-D$_{d-}$: GRU-D without time decay mechanism.
    \item \textit{CDNet}$_{+}$: \textit{CDNet} with a time decay mechanism \cite{che2018recurrent}.
\end{itemize}

The outputs of Mean, KNN, MICE, Simple, BRNN, and CATSI are fed into standard GRU to make mortality risk predictions.


\subsection{Implementation Details \& Evaluation Metrics}
We perform all the baselines and \textit{CDNet} with Python v3.9.7. We employ the following libraries: fancyimpute for KNN and MICE and PyTorch for the rest of the methods. For each task, we randomly split the datasets into training, validation, and testing sets in a 70:15:15 ratio. The validation set is used to select the best values of parameters. Training and evaluations were performed on A40 GPU from NVIDIA with 48GB of memory. We repeat all the methods ten times and report the average performance.

We use class weight in CrossEntropyLoss for a highly imbalanced dataset. This is achieved by placing an argument called 'weight' on the CrossEntropyLoss function (PyTorch).

We assess performance using the area under the receiver operating characteristic curve (AUROC) and the area under the precision-recall curve (AUPRC).

\subsection{Comparison with baselines}
\begin{table*}[htbp]
  \centering
  \caption{Performance of baselines and our approaches on in-hospital mortality prediction.}
    \begin{tabular}{ccccccc}
    \toprule
    MIMIC-III/Mortality Prediction & \multicolumn{2}{c}{24 hours after ICU admission} & \multicolumn{2}{c}{36 hours after ICU admission} & \multicolumn{2}{c}{48 hours after ICU admission} \\
    \midrule
    Metrics & AUROC & AUPRC & AUROC & AUPRC & AUROC & AUPRC \\
    \midrule
    Mean & 0.6780(0.017) & 0.2283(0.017) & 0.6821(0.016) & 0.2322(0.015) &	0.6816(0.016) & 0.2314(0.015) \\
    KNN & 0.7122(0.016) & 0.2498(0.022) & 0.7057(0.015) & 0.2423(0.019) & 0.7086(0.013) & 0.2464(0.019) \\
    MICE & 0.7089(0.019) & 0.2582(0.024) & 0.7113(0.020) & 0.2551(0.023) & 0.7058(0.018) & 0.2435(0.019) \\
    Simple & 0.6821(0.012) & 0.2315(0.010) & 0.6806(0.012) & 0.2307(0.010)	& 0.6791(0.013) & 0.2279(0.012) \\
    BRNN  & 0.6735(0.011) & 0.2037(0.012) & 0.6704(0.010) & 0.2023(0.012)	& 0.6732(0.011) & 0.2051(0.014) \\
    CATSI & 0.7042(0.011) & 0.2373(0.012) & 0.7024(0.013) & 0.2343(0.015)	& 0.7057(0.012) & 0.2379(0.012) \\
    BRITS & 0.7463(0.010) &  0.2880(0.016) & 0.7445(0.009) & 0.2856(0.016)	 & 0.7447(0.009) &  0.2879(0.016) \\
    GRU-D & 0.7323(0.012) & 0.2821(0.014) & 0.7235(0.012) & 0.2679(0.015) & 0.7285(0.011) & 0.2763(0.015) \\
    GRU-D$_{d-}$ & 0.7137(0.011) & 0.2689(0.016) & 0.7342(0.015) & 0.2624(0.015) & 0.7244(0.011) & 0.2673(0.014) \\
    \textit{CDNet}$_{\alpha}$ & 0.7536(0.008) & 0.3252(0.016) & 0.7502(0.011) & 0.3031(0.015) & 0.7546(0.008) & 0.2994(0.014) \\
    \textit{CDNet}$_{\beta}$ & 0.7557(0.013) & 0.3402(0.014) & 0.7538(0.010) & 0.3404(0.017) & 0.7543(0.012) & 0.3413(0.020) \\
    \textit{CDNet} & \textbf{0.7712(0.011)} & \textbf{0.3497(0.014)} & \textbf{0.7675(0.012)} & \textbf{0.3443(0.017)} & \textbf{0.7673(0.013)} & \textbf{0.3526(0.014)} \\
    \textit{CDNet}$_{+}$ & 0.7588(0.019) & 0.3286(0.017) & 0.7506(0.020) & 0.3166(0.017) & 0.7529(0.018) & 0.3177(0.015) \\
    \bottomrule
    \end{tabular}%
  \label{tab:addlabel}%
\end{table*}%
Table II lists the results of in-hospital mortality prediction based on the clinical times series data from the first 24, 36, and 48 hours after ICU admission, respectively. These results suggest that \textit{CDNet} significantly and consistently outperforms other baseline methods. Comparing the two results GRU-D$_{d-}$ and GRU-D in Table II (24 hours after ICU admission), it can be seen that the use of the time decay mechanism achieves a performance improvement in AUROC by 1.86\% and in AUPRC by 1.32\%. Interestingly, it can be seen from the data in Table II that GRU-D$_{d-}$ outperforms GRU-D in terms of AUROC by 1.07\% (36 hours after ICU admission). In addition, significant reductions in prediction performance of \textit{CDNet}$_{+}$ (\textit{CDNet} with a time decay mechanism) are observed compared with \textit{CDNet}. Taken together, these results suggest that there is high inconsistency on the effectiveness of the time decay mechanism.


\subsection{Ablation study}
We now proceed to examine the effectiveness of different components of our \textit{CDNet}. To this end, we conduct an ablation study on the datasets. We present two variants of \textit{CDNet} as:
\begin{itemize}
    \item \textit{CDNet}$_{\alpha}$: \textit{CDNet} without MDN and RAN.
    \item \textit{CDNet}$_{\beta}$: \textit{CDNet} without RAN.
\end{itemize}

We present the ablation study results in Table II. We find that \textit{CDNet}$_{\beta}$ outperforms \textit{CDNet}$_{\alpha}$. Overall, \textit{CDNet}$_{\beta}$ achieved significant performance improvements in AUPRC. These results demonstrate the effectiveness of the MDN construction. According to these results, we can also infer that \textit{CDNet}$_{\beta}$ is more concerned with the balance of classification than \textit{CDNet}$_{\alpha}$. The superior performance of \textit{CDNet} than \textit{CDNet}$_{\beta}$ verifies the efficacy of RAN, in achieving performance improvements in AUROC and AUPRC.

\subsection{Case study: Regularized Attention Network (RAN) Analysis}
Fig. 2 and Fig. 3 present the results of two patient journeys (two random examples) obtained from the RAN analysis. The boxes with attention scores are imputed values. The larger the attention scores, the more reliable the imputed values. The attention scores ranging from $0.0 \sim 1.0$ were calculated by the RAN. The RAN takes into consideration the entire patient journey, but the images are understandably truncated for visibility. We take the first 20 records of the two patient journeys as an example for detailed discussion. Between Fig. 2 and Fig. 3, there is a significant difference between the degree of missingness of the two patient journeys. We can observe that less reliable imputed values are assigned lower weights in the two patient journeys. These results suggest that RAN not only can handle the different degrees of missingness of patient journey data but also has fine-grained robustness at the feature level of patient journey data.
\begin{figure}[!htb]
        \centering
        \includegraphics[width = 1.0\linewidth]{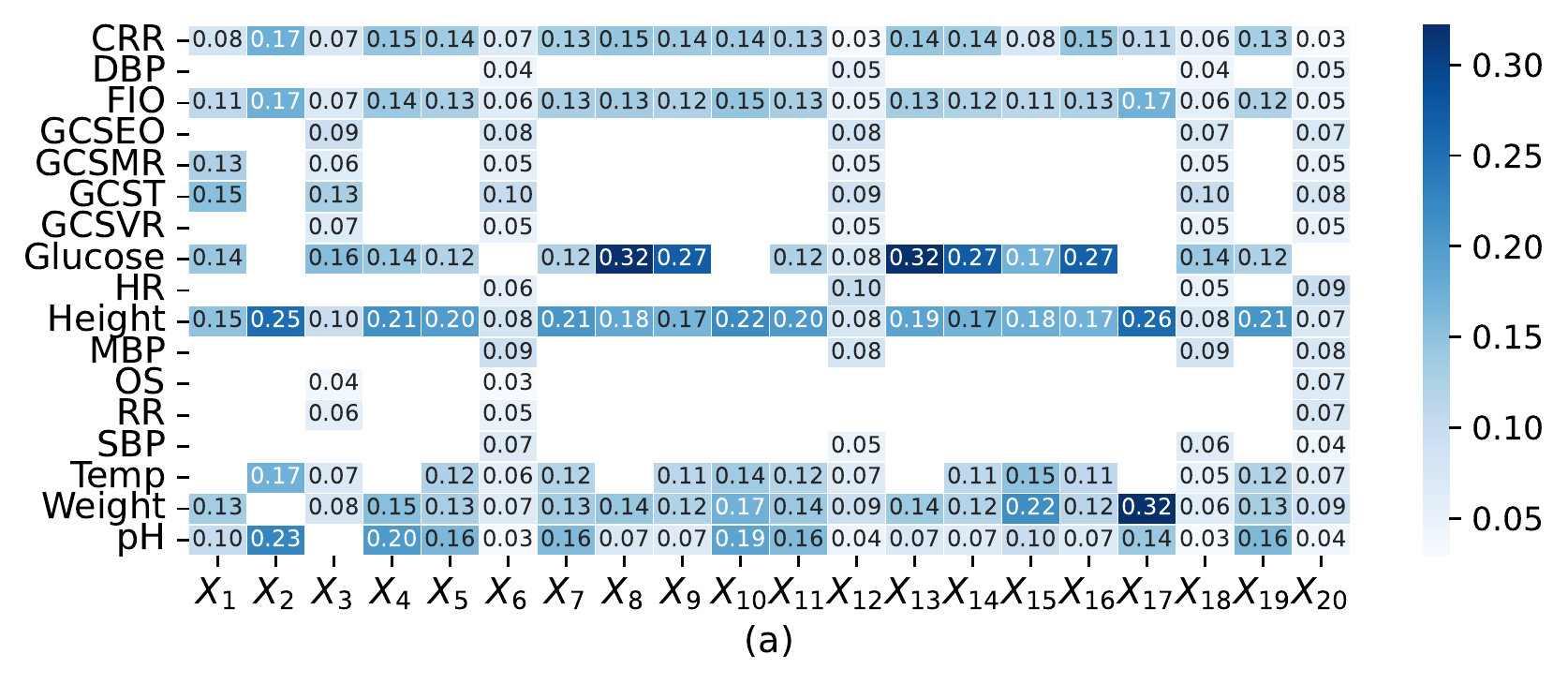}
        \caption{\textbf{Result of Patient A. }}
        \label{fig:RAN1}
\end{figure}
\begin{figure}[!htb]
        \centering
        \includegraphics[width = 1.0\linewidth]{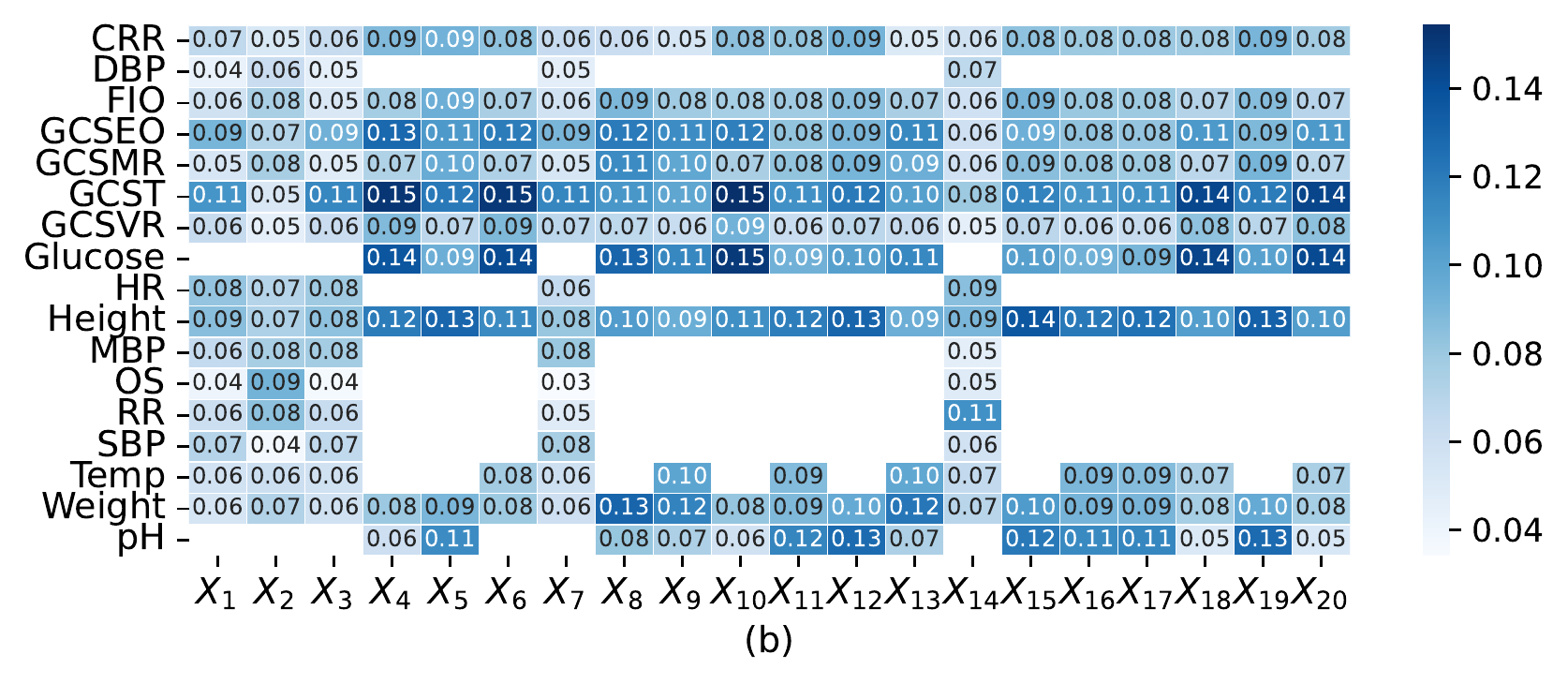}
        \caption{\textbf{Result of Patient B. }}
        \label{fig:RAN2}
\end{figure}

\subsection{Case study: Uncertainty Analysis}
Fig. 4 shows the results obtained from the epistemic uncertainty estimation of four patient journeys (four random examples). Each subgraph contains two predicted probability distributions of a patient journey, where dodger blue and dark orange histograms are derived from FFN ensembles and MDN, respectively. The MDN used here is described in the subsection Risk Prediction. For MDN, we set the mixture components to 100 (K = 100). In terms of a patient journey, these 100 components would produce 100 prediction results, so that epistemic uncertainty of the prediction model can be quantified. For FFN ensembles, we replace the MDN with 100 FNNs that have different random seeds. The more the two discrete distributions (histograms) overlap, the better the ability of the model to capture epistemic uncertainty. From the data in Fig. 4, it is apparent that there are many overlaps between the two discrete distributions in each subgraph. These results suggest that our method is able to capture epistemic uncertainty similar to that of FFN ensembles.
\begin{figure*}[!htb]
        \centering
        \includegraphics[width = 1.0\textwidth]{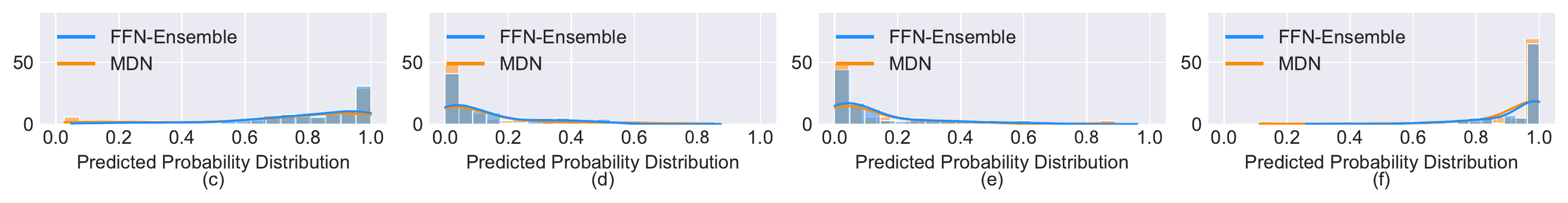}
        \caption{\textbf{Predicted probability distribution of MDN (our method) and FFN-ensemble.}}
        \label{fig:Uncertainty1}
\end{figure*}
\begin{figure*}[!htb]
        \centering
        \includegraphics[width = 1.0\linewidth]{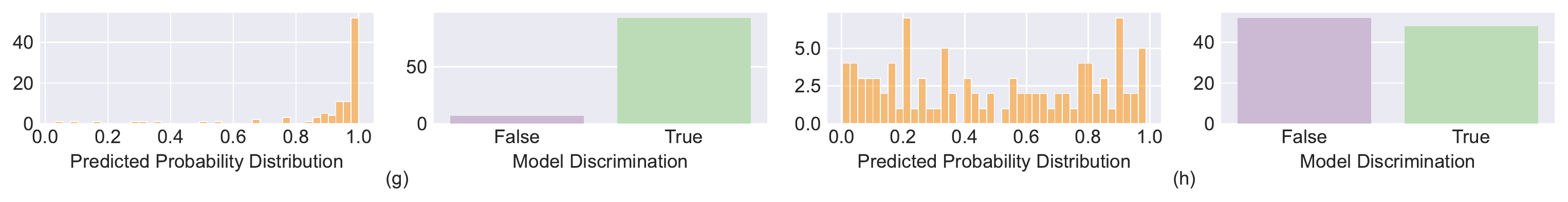}
        \caption{\textbf{Epistemic uncertainty analysis. Two examples of the predicted probability distribution on the mortality prediction task.}}
        \label{fig:Uncertainty2}
\end{figure*}
\begin{figure*}[!htb]
        \centering
        \includegraphics[width = 1.0\linewidth]{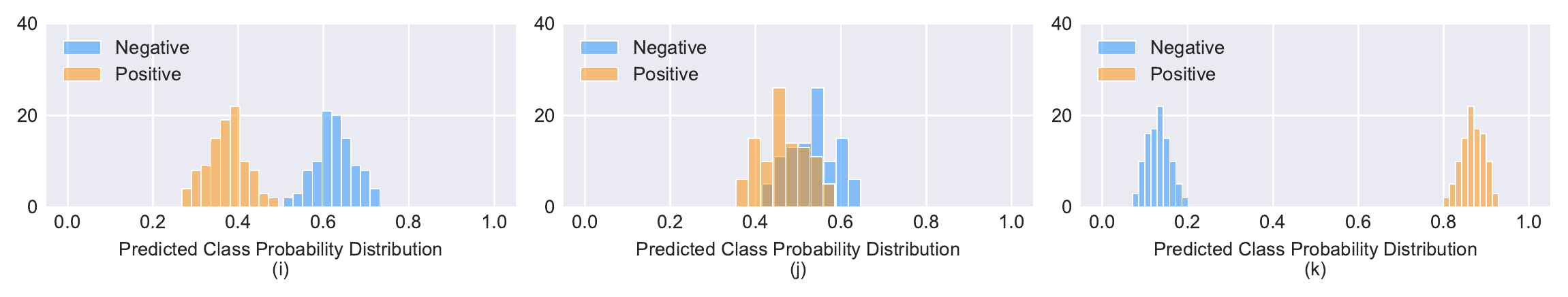}
        \caption{\textbf{Aleatoric uncertainty analysis. Three examples of the predicted probability distribution on the mortality prediction task.}}
        \label{fig:Uncertainty3}
\end{figure*}

Fig. 5 shows the results obtained from the epistemic uncertainty analysis of two patient journeys (two random examples). Fig. 5g (left) and Fig. 5h (left) compare the prediction results obtained from MDN. Fig. 5g (right) and Fig. 5h (right) show the corresponding model discriminations, with 'True' (likely to die) corresponding to a prediction of mortality and 'False' corresponding to the opposite (unlikely to die). From the patient in Fig. 5g, the agreement among ensemble members of MDN about 'True' is high. In contrast, there is high disagreement for another patient in Fig. 5h due to epistemic uncertainty.

Fig. 6 shows the results obtained from the aleatoric uncertainty analysis of three patient journeys (three random examples). Each subgraph contains two predicted class probability distributions of a patient journey, where dodger blue and dark orange histograms represent negative and positive classes, respectively, derived from MDN predictions. We augment MDN with 100 Gaussian noises to generate 100 class probability distributions for each of the two patient journeys. From the data in Fig. 6i and Fig. 6k, we can see that the histograms corresponding to the probability distributions of the two predicted classes do not overlap. The results suggest that aleatoric uncertainty had less impact on mortality predictions in these two cases. In addition, the negative class (unlikely to die) in Fig. 6i reported significantly higher probability than the other group. Similarly, the positive class (likely to die) in Fig. 6k reported significantly higher probability than the other group. As can be seen from the data in Fig. 6j, there is a large overlap between the histograms corresponding to the probability distributions of the two predicted classes. Thus, aleatoric uncertainty has a more significant impact on the mortality predictions of this patient. Although the result of model discrimination may be negative, this prediction should be taken with caution.

\section{CONCLUSION}
In this paper, we have presented a novel method of integrated training and regularizing a deep learning model with the aim of predicting patient health risk using EHRs with a large number of missing values. We validated the proposed model (\textit{CDNet}) on mortality prediction tasks using the MIMIC-III dataset that has a large degree of missingness in the input. Extensive experimental results demonstrated that \textit{CDNet} significantly outperformed existing methods. The ablation experiments proved that regularizing imputed values is a key factor for performance improvements. Further analysis of prediction uncertainty proved that our model could capture both aleatoric and epistemic uncertainties, which allows model users to know how reliable the results are.


\begin{thebibliography}{10}
\providecommand{\url}[1]{#1}
\csname url@samestyle\endcsname
\providecommand{\newblock}{\relax}
\providecommand{\bibinfo}[2]{#2}
\providecommand{\BIBentrySTDinterwordspacing}{\spaceskip=0pt\relax}
\providecommand{\BIBentryALTinterwordstretchfactor}{4}
\providecommand{\BIBentryALTinterwordspacing}{\spaceskip=\fontdimen2\font plus
\BIBentryALTinterwordstretchfactor\fontdimen3\font minus
  \fontdimen4\font\relax}
\providecommand{\BIBforeignlanguage}[2]{{%
\expandafter\ifx\csname l@#1\endcsname\relax
\typeout{** WARNING: IEEEtran.bst: No hyphenation pattern has been}%
\typeout{** loaded for the language `#1'. Using the pattern for}%
\typeout{** the default language instead.}%
\else
\language=\csname l@#1\endcsname
\fi
#2}}
\providecommand{\BIBdecl}{\relax}
\BIBdecl

\bibitem{dusenberry2020analyzing}
M.~W. Dusenberry, D.~Tran, E.~Choi, J.~Kemp, J.~Nixon, G.~Jerfel, K.~Heller,
  and A.~M. Dai, ``Analyzing the role of model uncertainty for electronic
  health records,'' in \emph{Proceedings of the ACM Conference on Health,
  Inference, and Learning}, 2020, pp. 204--213.

\bibitem{zhang2020inprem}
X.~Zhang, B.~Qian, S.~Cao, Y.~Li, H.~Chen, Y.~Zheng, and I.~Davidson, ``Inprem:
  An interpretable and trustworthy predictive model for healthcare,'' in
  \emph{Proceedings of the 26th ACM SIGKDD International Conference on
  Knowledge Discovery \& Data Mining}, 2020, pp. 450--460.

\bibitem{chen2021unite}
C.~Chen, J.~Liang, F.~Ma, L.~Glass, J.~Sun, and C.~Xiao, ``Unite:
  Uncertainty-based health risk prediction leveraging multi-sourced data,'' in
  \emph{Proceedings of the Web Conference 2021}, 2021, pp. 217--226.

\bibitem{johnson2016mimic}
A.~E. Johnson, T.~J. Pollard, L.~Shen, H.~L. Li-Wei, M.~Feng, M.~Ghassemi,
  B.~Moody, P.~Szolovits, L.~A. Celi, and R.~G. Mark, ``Mimic-iii, a freely
  accessible critical care database,'' \emph{Scientific data}, vol.~3, no.~1,
  pp. 1--9, 2016.

\bibitem{tan2020data}
Q.~Tan, M.~Ye, B.~Yang, S.~Liu, A.~J. Ma, T.~C.-F. Yip, G.~L.-H. Wong, and
  P.~Yuen, ``Data-gru: Dual-attention time-aware gated recurrent unit for
  irregular multivariate time series,'' in \emph{Proceedings of the AAAI
  Conference on Artificial Intelligence}, vol.~34, no.~01, 2020, pp. 930--937.

\bibitem{khan2020machine}
A.~I. Khan and S.~Al-Habsi, ``Machine learning in computer vision,''
  \emph{Procedia Computer Science}, vol. 167, pp. 1444--1451, 2020.

\bibitem{singh2017machine}
S.~P. Singh, A.~Kumar, H.~Darbari, L.~Singh, A.~Rastogi, and S.~Jain, ``Machine
  translation using deep learning: An overview,'' in \emph{2017 international
  conference on computer, communications and electronics (comptelix)}.\hskip
  1em plus 0.5em minus 0.4em\relax IEEE, 2017, pp. 162--167.

\bibitem{tarca2007machine}
A.~L. Tarca, V.~J. Carey, X.-w. Chen, R.~Romero, and S.~Dr{\u{a}}ghici,
  ``Machine learning and its applications to biology,'' \emph{PLoS
  computational biology}, vol.~3, no.~6, p. e116, 2007.

\bibitem{bertsimas2021imputation}
D.~Bertsimas, A.~Orfanoudaki, and C.~Pawlowski, ``Imputation of clinical
  covariates in time series,'' \emph{Machine Learning}, vol. 110, no.~1, pp.
  185--248, 2021.

\bibitem{welch2014evaluation}
C.~A. Welch, I.~Petersen, J.~W. Bartlett, I.~R. White, L.~Marston, R.~W.
  Morris, I.~Nazareth, K.~Walters, and J.~Carpenter, ``Evaluation of two-fold
  fully conditional specification multiple imputation for longitudinal
  electronic health record data,'' \emph{Statistics in medicine}, vol.~33,
  no.~21, pp. 3725--3737, 2014.

\bibitem{bounthavong2015approach}
M.~Bounthavong, J.~H. Watanabe, and K.~M. Sullivan, ``Approach to addressing
  missing data for electronic medical records and pharmacy claims data
  research,'' \emph{Pharmacotherapy: The Journal of Human Pharmacology and Drug
  Therapy}, vol.~35, no.~4, pp. 380--387, 2015.

\bibitem{harutyunyan2019multitask}
H.~Harutyunyan, H.~Khachatrian, D.~C. Kale, G.~Ver~Steeg, and A.~Galstyan,
  ``Multitask learning and benchmarking with clinical time series data,''
  \emph{Scientific data}, vol.~6, no.~1, pp. 1--18, 2019.

\bibitem{sheikhalishahi2019benchmarking}
S.~Sheikhalishahi, V.~Balaraman, and V.~Osmani, ``Benchmarking machine learning
  models on eicu critical care dataset,'' \emph{arXiv preprint
  arXiv:1910.00964}, 2019.

\bibitem{che2018recurrent}
Z.~Che, S.~Purushotham, K.~Cho, D.~Sontag, and Y.~Liu, ``Recurrent neural
  networks for multivariate time series with missing values,'' \emph{Scientific
  reports}, vol.~8, no.~1, pp. 1--12, 2018.

\bibitem{cho2014learning}
K.~Cho, B.~Van~Merri{\"e}nboer, C.~Gulcehre, D.~Bahdanau, F.~Bougares,
  H.~Schwenk, and Y.~Bengio, ``Learning phrase representations using rnn
  encoder-decoder for statistical machine translation,'' \emph{arXiv preprint
  arXiv:1406.1078}, 2014.

\bibitem{abdar2021review}
M.~Abdar, F.~Pourpanah, S.~Hussain, D.~Rezazadegan, L.~Liu, M.~Ghavamzadeh,
  P.~Fieguth, X.~Cao, A.~Khosravi, U.~R. Acharya \emph{et~al.}, ``A review of
  uncertainty quantification in deep learning: Techniques, applications and
  challenges,'' \emph{Information Fusion}, vol.~76, pp. 243--297, 2021.

\bibitem{hullermeier2021aleatoric}
E.~H{\"u}llermeier and W.~Waegeman, ``Aleatoric and epistemic uncertainty in
  machine learning: An introduction to concepts and methods,'' \emph{Machine
  Learning}, vol. 110, no.~3, pp. 457--506, 2021.

\bibitem{li2016aleatory}
G.~Li, Z.~Lu, L.~Li, and B.~Ren, ``Aleatory and epistemic uncertainties
  analysis based on non-probabilistic reliability and its kriging solution,''
  \emph{Applied Mathematical Modelling}, vol.~40, no. 9-10, pp. 5703--5716,
  2016.

\bibitem{kendall2017uncertainties}
A.~Kendall and Y.~Gal, ``What uncertainties do we need in bayesian deep
  learning for computer vision?'' \emph{Advances in neural information
  processing systems}, vol.~30, 2017.

\bibitem{rohekar2018constructing}
R.~Y. Rohekar, S.~Nisimov, Y.~Gurwicz, G.~Koren, and G.~Novik, ``Constructing
  deep neural networks by bayesian network structure learning,'' \emph{Advances
  in Neural Information Processing Systems}, vol.~31, 2018.

\bibitem{hoffman2013stochastic}
M.~D. Hoffman, D.~M. Blei, C.~Wang, and J.~Paisley, ``Stochastic variational
  inference,'' \emph{Journal of Machine Learning Research}, 2013.

\bibitem{domke2018importance}
J.~Domke and D.~R. Sheldon, ``Importance weighting and variational inference,''
  \emph{Advances in neural information processing systems}, vol.~31, 2018.

\bibitem{dhaka2020robust}
A.~K. Dhaka, A.~Catalina, M.~R. Andersen, M.~Magnusson, J.~Huggins, and
  A.~Vehtari, ``Robust, accurate stochastic optimization for variational
  inference,'' \emph{Advances in Neural Information Processing Systems},
  vol.~33, pp. 10\,961--10\,973, 2020.

\bibitem{acerbi2018variational}
L.~Acerbi, ``Variational bayesian monte carlo,'' \emph{Advances in Neural
  Information Processing Systems}, vol.~31, 2018.

\bibitem{gal2016dropout}
Y.~Gal and Z.~Ghahramani, ``Dropout as a bayesian approximation: Representing
  model uncertainty in deep learning,'' in \emph{international conference on
  machine learning}.\hskip 1em plus 0.5em minus 0.4em\relax PMLR, 2016, pp.
  1050--1059.

\bibitem{guillaumes2017mixture}
A.~B. Guillaumes, ``Mixture density networks for distribution and uncertainty
  estimation,'' Ph.D. dissertation, Universitat Polit{\`e}cnica de Catalunya.
  Facultat d'Inform{\`a}tica de Barcelona, 2017.

\bibitem{van2011mice}
S.~Van~Buuren and K.~Groothuis-Oudshoorn, ``mice: Multivariate imputation by
  chained equations in r,'' \emph{Journal of statistical software}, vol.~45,
  pp. 1--67, 2011.

\bibitem{suo2019recurrent}
Q.~Suo, L.~Yao, G.~Xun, J.~Sun, and A.~Zhang, ``Recurrent imputation for
  multivariate time series with missing values,'' in \emph{2019 IEEE
  International Conference on Healthcare Informatics (ICHI)}.\hskip 1em plus
  0.5em minus 0.4em\relax IEEE, 2019, pp. 1--3.

\bibitem{yin2019context}
K.~Yin and W.~K. Cheung, ``Context-aware imputation for clinical time series,''
  in \emph{2019 IEEE International Conference on Healthcare Informatics
  (ICHI)}.\hskip 1em plus 0.5em minus 0.4em\relax IEEE, 2019, pp. 1--3.

\bibitem{cao2018brits}
W.~Cao, D.~Wang, J.~Li, H.~Zhou, L.~Li, and Y.~Li, ``Brits: Bidirectional
  recurrent imputation for time series,'' \emph{Advances in neural information
  processing systems}, vol.~31, 2018.

\end{thebibliography}
\end{document}